%% file: paper.tex
\documentclass[letterpaper, 10 pt, conference]{ieeeconf}  % Comment this line out if you need a4paper

\usepackage{bm}
\usepackage[noadjust]{cite}
\usepackage{color}
\usepackage{amsmath}
\usepackage{xr}
\usepackage{amssymb}
\usepackage{float}
\usepackage{algorithm}
\usepackage{empheq}
\let\labelindent\relax
\usepackage[shortlabels]{enumitem}
\usepackage{siunitx}
\usepackage{multirow}
\usepackage[labelformat=simple]{subcaption}

\usepackage{algpseudocode}
\algnewcommand{\LineComment}[1]{\State \(\triangleright\) #1}

\IEEEoverridecommandlockouts                              % This command is only needed if 
\title{\LARGE \bf An Interval Branch-and-Bound-Based Inverse Kinematics Algorithm for Complex Robotic Manipulation Tasks}
\author{
Yajue Yang, Zeqing Zhang, Yuanqing Wu, Jia Pan
}

\begin{document}

\maketitle
\thispagestyle{empty}
\pagestyle{empty}

%%%%%%%%%%%%%%%%%%%%%%%%%%%%%%%%%%%%%%%%%%%%%%%%%%%%%%%%%%%%%%%%%%%%%%%%%%%%%%%%
\begin{abstract}
Even if the agent has made great progress in manipulation, the general inverse kinematics (IK) problem of a serial manipulator, i.e., the acquisition of the \emph{self-motion manifold} (SMM) of all admissible joint angles for a desired end-effector pose, still plays a fundamental role in robotic manipulation tasks. To efficiently solve the generalized IK, this paper proposes an interval branch-and-bound approach augmented with a fast numerical IK-solver-enabled search heuristic. Compared with independent solutions generated by sampling-based methods, our approach produces patches of neighboring solutions, providing richer information about the inherent geometry of the SMM for optimal planning and other applications. It can also be used at any time to obtain solutions with sub-optimal resolution for applications within a limited period. The performance of our approach is verified by numerical experiments on both non-redundant and redundant manipulators.
\end{abstract}

%%%%%%%%%%%%%%%%%%%%%%
% Outline
% 1. Introduction
% 	a) Self motion (definition, applications)
%   b) Approaches to self motion (random sampling, bnb)
%   c) Contribution in this paper
%
% 2. Related works
%   a) Problem formulation (rotation matrix)
%   b) BnB framework (contractor and bisector)
%
% 3. Algorithm
%
% 4. Experiments
%
% 5. Conclusion (future work)
%%%%%%%%%%%%%%%%%%%%%%

\input{intro}

\input{bnb_overview}
\input{prob_formulation}
\input{heur_manif}

\input{experiments}
\input{conclusion}

{\small
\bibliographystyle{IEEEtran}
\bibliography{reference}
}

\end{document}

%% file: intro.tex
%%%%%%%%%%%%%%%%%%%%%%
% Outline
% 1. Introduction
% 	a) Inverse kinematics of redundant manipulators ---> self-motion (definition, applications)
%   b) Approaches to self motion (analytical random sampling, bnb)
%   c) Contribution in this paper and the structure
%%%%%%%%%%%%%%%%%%%%%%

\externaldocument{bnb_overview}
\externaldocument{prob_formulation}
\externaldocument{heur_manif}
\externaldocument{experiments}
\externaldocument{conclusion}

\section{Introduction}
\label{sec: intro}
% introduction to IK
When planning a kinematically redundant robot arm that has more degrees of freedom (DoF) than those required to perform a given manipulation task in a cluttered environment, it is of imperative importance to take advantage of its redundant freedoms for collision avoidance, continuous joint/end-effector trajectory generation and maximization of kineto-dynamic performance~\cite{chiaverini1997singularity, hauser2018global}.

\begin{figure*}
    \centering
    \begin{subfigure}[b]{0.3\textwidth}
        \includegraphics[width=\textwidth]{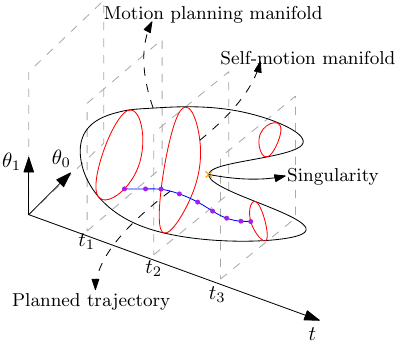}
        \caption{The EE path tracking problem}
        \label{fig: ep_illustration}
    \end{subfigure}
    \begin{subfigure}[b]{0.3\textwidth}
        \includegraphics[width=\textwidth]{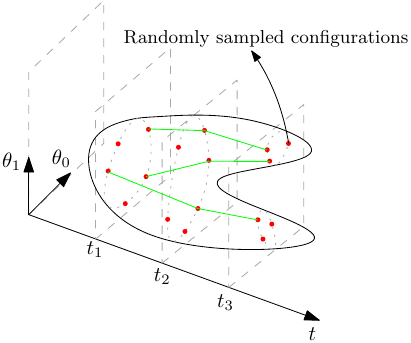}
        \caption{Randomized sampling-based approach}
        \label{fig: ep_rs}
    \end{subfigure}
    \begin{subfigure}[b]{0.3\textwidth}
        \includegraphics[width=\textwidth]{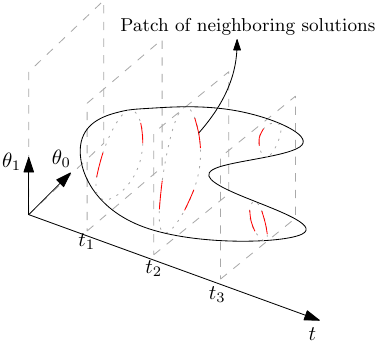}
        \caption{SMM exploration with the BnB}
        \label{fig: ep_bnb}
    \end{subfigure}
    \vspace{-5pt}
    \caption{Illustration of the EE path tracking problem and related techniques.}
    \label{fig:animals}    
\end{figure*}

For example, consider generation of a globally optimal trajectory (maximizing certain criteria) along a given end-effector (EE) path of a redundant manipulator. Denoting by $n$ and $m$ the DoF and task space dimension of the redundant robot, a given EE pose can in general be reached by setting its joint angle vector to lie on any of a finite number of disjointed submanifolds (with boundary) of dimension $n-m$  in the joint configuration space; they shall be collectively referred to as the \emph{self-motion manifold} (SMM)~\cite{burdick1989inverse}. More generally, the one-parameter family of SMMs along a specified EE trajectory constitutes the feasible search space for the optimal path, and will be referred to as \emph{self-motion domain} (SMD). The topology change of the slices (members of the one-parameter family of SMMs) of the SMD , as illustrated in Fig.~\ref{fig: ep_illustration}, may prevent the robot from continuously traversing the full EE path~\cite{luck1997self, peidro2018method}. Therefore, it is arguably true that globally optimal collision-free path planning can only be achieved with any guarantee if (at least partial) topological information concerning branching and connectedness of the SMM is available. This in turn requires a comprehensive (implicit and/or explicit) characterization of the SMD, which we refer to as the \emph{generalized} inverse kinematics (IK) problem in this paper.

\cite{hauser2018global} and~\cite{oriolo2005motion} proposed to construct a tree/roadmap in which nodes represent random joint configurations on SMMs for discretized EE poses, and edges connect transferable nodes associated with two adjacent EE poses, illustrated in Fig.~\ref{fig: ep_rs}. An numerical instantaneous IK solver is utilized to quickly find isolated solutions on SMMs with random initial joint values. However, this approach can not guarantee \emph{completeness}, \emph{i.e.}, finding all feasible IK solutions. On the other hand, its random nature does not lead to a topological characterization of the SMMs. To achieve completeness, \cite{luck1997self} and \cite{peidro2018method} resort to first discretizing a selected set of redundant joints and then solving, for each redundant joint value, a non-redundant IK problem for the remaining joints. Such methods are intuitive to follow, but can not generate dense uniformly spaced samples on the SMM due to singularities; they require extra procedures to ensure a proper coverage of the entire SMD.

% \captionsetup[subfloat]{listofformat=parens}

By considering the general IK problem as a constraint satisfaction problems (CSPs), i.e., search of feasible argument values that satisfy a set of constraints, one can utilize the interval branch and bound (BnB) approach, a general global framework for solving CSPs, to completely enclose all IK solutions within sufficiently small interval boxes~\cite{rao1998inverse,castellet1998algorithm,merlet2010interval}. A second advantage of the interval-based algorithms is their distinct capability of dealing with system parameter uncertainties ubiquitous to all robotic systems~\cite{merlet2010interval,araya2016interval,castellet1998algorithm}. The BnB approach is also easy to implement in the sense that it only requires a proper constraint formulation of the problem at hand. In the case of the IK problem, the BnB approach is capable of handling different types of EE pose targets (\emph{e.g.}, fully fixed $6$D pose and the pose with a certain position and $Z$-axis direction); it could also handle other constraints, such as joint limits.
% Fundamentally, the BnB approach has potential for broad applications since what it needs is just a proper constraint formulation. To be specific, the BnB approach is flexible in various types of EE pose targets (\emph{e.g.}, fully fixed $6$D pose and the pose with a certain position and $Z$-axis direction); and could easily handle other constraints, such as joint limits.

% proposed method
This paper aims to exploit the sophistication and generality of the BnB approach for complete and efficient solutions of the generalized IK problem. To address the inherent efficiency issue of the BnB approach~\cite{chenouard2009search}, we propose, on the one hand, to accelerate the BnB approach by implementing a fast instantaneous IK solver in the framework to provide search heuristics, and, on the other hand, use them in an anytime fashion. Results of our algorithm shows that the BnB framework and the local IK solver can efficiently complement each other: while the local IK solver can rapidly find an isolated solution using random sampled initial joint values, the BnB framework further explores the portion of the SMM emanating from that solution, and eventually traverse the entire search space prescribing the SMM. In comparison to infinitesimal IK solvers that generates one solution at a time, the proposed algorithm outputs a patch of neighboring solutions in an anytime fashion, which arguably provides richer topological information of the SMM for optimal motion planning and other applications as depicted in Fig.~\ref{fig: ep_bnb}.

The paper is organized as follows. Section~\ref{sec: bnb_overview} gives an overview of the framework of interval BnB algorithms.
Section~\ref{sec: prob_formulation} reformulates the IK problem for implementing the BnB algorithm. Section~\ref{sec: heur_manif} proposes an interval BnB algorithm with search heuristics, in which a SMM paving strategy tailored for $1$ DoF redundancy is incorporated. Results of experiments are illustrated and analyzed in Section~\ref{sec: experiments}. Finally, the paper is concluded in Section~\ref{sec: conclusion}.

%% file: bnb_overview.tex
\vspace{-5pt}
\begin{algorithm}
\caption{Interval BnB Framework}
\begin{algorithmic}[1]
\Require initial box $[x] \in \mathbb{IR}^n$, minimum box size $\epsilon > 0$
\Ensure the set of solutions $\mathcal{S} = \{[x_1], [x_2], \dots, [x_n]\}$
\State $\mathcal{S} \gets \emptyset$
\State unexplored box buffer $ \mathcal{L} \gets \{[x]\}$
\While {$\mathcal{L} \neq \emptyset$}
\State $[x] \gets \textsc{Extract}(\mathcal{L})$ \Comment{Select a box from $\mathcal{L}$}
\State $[x] \gets \textsc{Contract}([x])$
\If {$[x] \neq \emptyset$}
\LineComment{If the box is small enough, insert it into $\mathcal{S}$.}
\If {$\textsc{Width}([x]) < \epsilon$}
\State $\mathcal{S} \gets [x] \cup \mathcal{S}$
\LineComment{If the box is still large, bisect it.}
\Else
\State $([x_1], [x_2]) \gets \textsc{Bisect}([x])$
\State $\mathcal{L} \gets \mathcal{L} \cup ([x_1], [x_2])$
\EndIf
\EndIf
\EndWhile
\end{algorithmic}
\label{alg: framework}
\end{algorithm}
\vspace{-15pt}

\section{Overview of Interval BnB}
\label{sec: bnb_overview}

\begin{figure*}
\centering
\includegraphics[width=\linewidth]{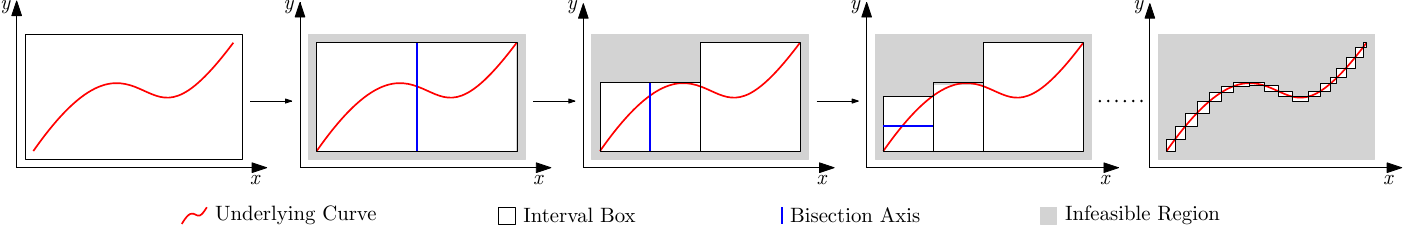}
\caption{Sequential splitting and shrinking of boxes enclosing all solutions of a nonlinear equation using interval BnB algorithm.}
\label{fig: bnb_example}
\end{figure*}

Consider a continuous CSP with a bounded domain, the interval BnB approach comprises of an iterative sequence of steps, alternating between \emph{contraction} (bounding) and \emph{bisection} (branching) to envelop all feasible solutions, until the size of all bounding boxes reach a minimum value $\epsilon$. Fig.~\ref{fig: bnb_example} depicts this interlacing process with a $2$D example; the generic interval BnB framework is illustrated in Algorithm~\ref{alg: framework}.

Let $\mathbb{IR}^n$ denote the set of interval vectors with components being a real-valued interval, a contraction operation associated with a set of constraints is defined as a mapping $\mathcal{C}: \mathbb{IR}^n \to \mathbb{IR}^n $ between two n-dimensional interval vectors~\cite{chabert2009contractor} such that: \[ \begin{cases}
\forall [x] \in \mathbb{IR}^n,\ \mathcal{C}([x]) \subseteq [x] \\
\forall x \in [x]\setminus \mathcal{C}([x]),\ \text{if the constraint is not satisfied.}
\end{cases} \]
The width of a interval vector (denoted with $\textsc{Width}([x])$) is the largest diameter of its components. More details about interval analysis can be found in~\cite{moore1979methods}. A contractor may comprise a sequence of different \emph{atomic} contractors, each taking the shrunk output of its predecessor as the input box. Contraction algorithms can often be classified into three types --- contractors based on interval arithmetic~\cite{hansen1992bounding}, contractors with constraint propagation~\cite{chabert2009contractor} and linear relaxation-based contractors~\cite{lebbah2005efficient}. Depending on the type of constraints and size of input box, the three types of contractors may differ significantly in effectivenss and efficiency. Therefore, an elaborate combination of atomic contractors can potentially optimize the overall performance of the contraction step. On the other hand, methods for the bisection step mainly differs in the selection of the component of the box to be split, such as Round-Robin, Largest-First and smear-based methods~\cite{araya2016interval}. Due to the absence of a suitable selection strategy (as far as we are aware), trial and error method is used to determine the suitable bisector(s).

%% file: prob_formulation.tex
% problem definition
\section{Problem Formulation}
\label{sec: prob_formulation}
The IK problem is defined as the inverse problem of the forward kinematics map of a robot manipulator:
\begin{equation}
\mathbf x=f(\boldsymbol\theta) 
\end{equation}
where $\boldsymbol\theta$ and $\mathbf x$ denote respectively the joint angle vector and EE pose. The forward kinematics map $f$ is usually represented by the product of exponential formula~\cite{murray2017mathematical}. Such formulation however leads to a low efficiency of the interval BnB algorithms, because the same joint variables may occur multiple times in one equation, and consequently slows down interval computations~\cite{merlet2010interval}. Fortunately, we can resort to intermediate variables proposed by Porta~\emph{et al.}~\cite{porta2009linear}, in which the kinematic constraints are formulated over rotation matrix components of each link; the positions of all links can be expressed in terms of their orientations. Consequently, the entire nonlinear constraint system comprises only three types of computationally light equations: linear, quadratic and bilinear. To finally solve the IK problem, the rotation matrix components are subsequently converted to the final variables $\bm{\theta}$. The rest of this section is devoted to details concerning the two steps.

\subsection{Kinematics Equations with Rotation Matrices}
It is well known that holonomically constrained multibody kinematics system can be characterized by a set of joint equations:
\begin{subequations}
\begin{empheq}[left=\empheqlbrace]{align}
\mathbf{r}_{i+1} + R_{i+1}\mathbf{p}_i^{\mathcal{F}_{i+1}} &= \mathbf{r}_{i} + R_{i}\mathbf{p}_i^{\mathcal{F}_{i}} \label{eq: joint_eq_1}\\
R_{i+1}\mathbf{d}_i^{\mathcal{F}_{i+1}} &= R_{i}\mathbf{d}_i^{\mathcal{F}_{i}}
\label{eq: joint_eq_2}
\end{empheq}
\end{subequations}
These equations describe relative allowable motions between two adjacent links indexed with $i$ and $i+1$. Here, $\mathcal{F} = (\mathbf{r}, R)$ denotes the pose of each link with respect to an inertial coordinate frame; $\mathbf{d}_i$ is the axis vector of joint $i$ that is fixed to the link $i$ and $\mathbf{p}_i$ is an arbitrary point on this axis. The superscripts such as in $\mathcal{F}_i$ stand for the body fixed link frame under which a point or vector is expressed. By recursively applying Eq.~\eqref{eq: joint_eq_1}, we have the following relationship between the pose of EE and other links:
\begin{equation}
\medmuskip=0mu
\thinmuskip=0mu
\thickmuskip=0mu
\mathbf{r}_{n+1} + R_{n+1}\mathbf{p}_{n}^{\mathcal{F}_{n+1}} = \mathbf{r}_{1} + R_{1}\mathbf{p}_1^{\mathcal{F}_{1}} + \sum_{j=2}^n R_j(\mathbf{p}_j - \mathbf{p}_{j-1})^{\mathcal{F}_j}
\label{eq: position_constraint}
\end{equation}
where $j = 1$ represents the base link.
In addition, each rotation matrix $R_i = (\mathbf{u}_i, \mathbf{v}_i, \mathbf{w}_i)$ must satisfy the orthogonality constraints for its validity:
\begin{equation}
||\mathbf{u}_i|| = 1,\ ||\mathbf{v}_i|| = 1,\ 
\mathbf{u}_i \cdot \mathbf{v}_i = 0,\ 
\mathbf{u}_i \times \mathbf{v}_i = \mathbf{w}_i.
\label{eq: rot_constraints}
\end{equation}
Finally, given a pose $(\mathbf{r}_{n+1}, R_{n+1})$, we can solve Eq.~\eqref{eq: joint_eq_2},~\eqref{eq: position_constraint} and~\eqref{eq: rot_constraints} for the remaining rotation matrices.

\subsection{Deriving Joint Values from Rotation Matrices}
As briefly mentioned in~\cite{porta2009linear}, joint angles can be derived from adjacency relationships between rotation matrices. We shall now explicitly formulate this procedure. After global link frame locations are calculated, the relative transformation $\mathcal{F}_{i, i+1}$ of the link $i+1$ with respect to link $i$ can be computed by:
\begin{equation}
\mathcal{F}_{i, i+1} = \mathcal{F}^{-1}_i\mathcal{F}_{i+1}.
\label{eq: rel_tf}
\end{equation}
Assuming that each joint only has one DoF, any rigid motion can be expressed by:
\begin{equation}
\mathcal{F}_{i, i+1} = e^{\hat{\xi}_i^{\mathcal{F}_i}\theta_i}\mathcal{F}_{i, i+1}(0),
\label{eq: exp_map}
\end{equation}
where $\hat{\xi}_i^{\mathcal F}$ denotes the twist coordinate of $i$th joint represented in $\mathcal F$~\cite{murray2017mathematical}, and $\mathcal{F}_{i, i+1}(0)$ is the initial relative transformation. Combining~\eqref{eq: rel_tf} and~\eqref{eq: exp_map}, it follows that
\begin{equation}
e^{\hat{\xi}_i^{\mathcal{F}_i}\theta_i} = \mathcal{F}_i\mathcal{F}_{i+1}\mathcal{F}_{i, i+1}(0)^{-1}.
\end{equation}
from which the joint angle $\theta_i$ can be easily computed~\cite{murray2017mathematical}.

%% file: heur_manif.tex
\begin{algorithm}[t]
\caption{BnB with Search Heuristics}
\begin{algorithmic}[1]
\Require initial rotation matrix elements box $[x]$, minimum box size $\epsilon > 0$, end time $t_e$
\Ensure the set of curves $\mathcal{S} = \{C_1, C_2, \dots, C_n\}$ where $C_i = \{[\theta_1], [\theta_2], \dots, [\theta_n]\}$ and $[\theta]$ is the joint interval
\State $\mathcal{S} \gets \emptyset$
\State unexplored box buffer $ \mathcal{L} \gets \{[x]\}$
\While {$\mathcal{L} \neq \emptyset$ \text{and} $t < t_e$}
\State (success, $\theta_s$) $\gets$ \textsc{CalcNewSol}($\mathcal{S}, n$)
\If {success is true}
\State \textsc{ExploreManif}($\theta_s$)
\Else
\State \textsc{SearchBuffer}($\mathcal{L}$)
\EndIf
\EndWhile
\end{algorithmic}
\label{alg: heur_manif}
\end{algorithm}
\vspace{-5pt}

\section{Algorithm Heuristic and Manifold Paving}
\label{sec: heur_manif}
The proposed algorithm relies on a synergy of the global BnB framework with numerical IK algorithms based on Newton-type methods and/or other nonlinear optimization techniques~\cite{beeson2015trac}. On the one hand, the local solver assists the BnB solver, in the early stage, ceaselessly obtaining solutions by directly specifying location of the search region, thereby enabling the anytime usage. For instance, in the EE tracking problem, patches of SMMs for different EE poses can be simultaneously generated in anytime and then be utilized to generate a sub-optimal trajectory. On the other hand, based on the specified isolated solution, the BnB solver could further expand a local solution patch or even elongate a self-motion curve, thereby providing topological information missing from the local IK solvers. In the case of $1$ DoF redundancy, we may also use a subroutine to pave the $1$D SMM to provide relevant information, such as the tangent directions at points on each curve and the number of disjointed curves, which is useful for early termination of the solving.

The complete algorithm is summarized in Algorithm~\ref{alg: heur_manif}. At the start of each iteration, the local IK solver attempts to acquire a new solution (Line 4), of which the success rate will gradually decrease due to gradual shrinking of unexplored solution regions. If a new solution is found, the BnB solver searches for the $\epsilon$-width box containing that solution and then explores the surrounding region on the SMM (Line 6). Otherwise, the BnB solver will examine one of the remaining boxes in the unexplored box list (Line 8). Every time an interval solution is acquired during \textsc{ExploreManif} and \textsc{SearchBuffer}, the solution will be categorized and arranged into different self-motion curves. Finally, the program terminates after the space has been thoroughly searched or when the allowable calculation time is over. The function modules are described in detail as follows.

\begin{figure}[t]
\includegraphics[width=\linewidth]{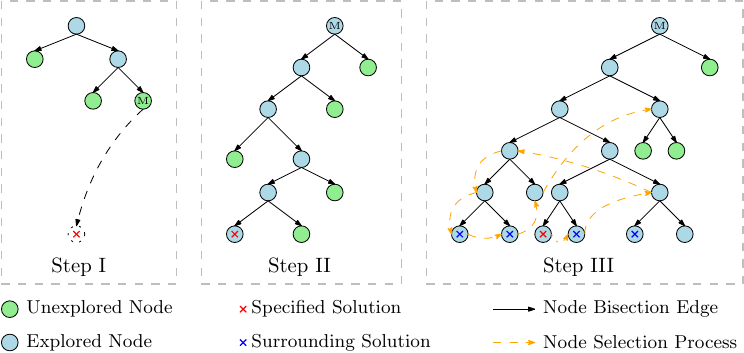}
\caption{Manifold exploration by the DFS. Step I finds the box M containing the specified solution in the BnB search tree. Step II searches for the $\epsilon$-width box in M by bisecting and contracting the sub-box containing the specified solution. Step III explores the surrounding region.}
\label{fig: expand_local_manif}
\vspace{-10pt}
\end{figure}

\begin{figure}
\includegraphics[width=\linewidth]{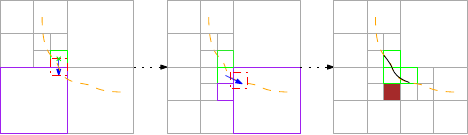}
\caption{Manifold exploration based on continuation. The curve to be paved is depicted with the dashed orange line. Gray grids represent nodes of the BnB search tree. At a solution box(the green grid), the extension direction (the blue arrow) is estimated, along which the box solution is slightly shifted to the red dashed box. Each of the unexplored intersecting boxes (purple) is then checked for the existence of adjacent solutions. Infeasible boxes are shown by the brown filled grid.}
\label{fig: expand_manif_cont}
\vspace{-10pt}
\end{figure}

Our BnB solver uses the following two strategies to find neighboring solution of a given one:
\begin{enumerate}[wide]
\item \textbf{Depth First Search (DFS):} This method fundamentally relies on the feature of the BnB search tree construction. As shown in Fig.~\ref{fig: expand_local_manif}, when a branch in the tree grows toward a specified solution, other leaf nodes on the same branch, \emph{i.e.}, the potential neighboring solutions, are pushed back into the unexplored box stack. Therefore, a DFS (last-in-first-out) facilitates a continuous exploration of the surrounding space.
\item \textbf{Manifold Continuation (MC):} In this method, the BnB solver searches solutions along a self-motion curve under the guidance of the tangential direction on the self-motion curve. Assuming that the interval box is small enough, this direction is approximated with the tangent vector at the middle point. More specifically, given a solution, the unexplored box that is near to the solution and in the direction of the curve's extension is selected for adjacent solutions. Since such box cannot be accurately specified, we resort to finding unexplored box candidates that intersect with the slightly shifted solution box (along the extension direction). Fig.~\ref{fig: expand_manif_cont} depicts a $2$D version of the continuation search process.
\end{enumerate}

\begin{figure}[t]
\centering
\includegraphics[width=\linewidth]{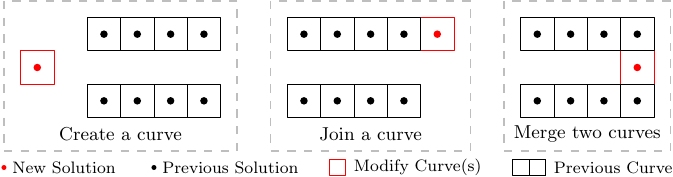}
\caption{Paving $1$D manifolds. }
\label{fig: curve_paving}
\vspace{-10pt}
\end{figure}

% pros and cons
Both methods have their own strengths and weaknesses and could be utilized according to application requirements. In general, the DFS is more efficient than the MC, because the box nodes to be examined in the DFS are usually smaller than those in the MC. However, the DFS might output a discontinuous patch of neighboring solutions, since no measure is taken to deliberately bisect nodes so that the sibling sub-nodes in the search tree are true neighbors on the SMM. On the contrary, the MC is designed to orderly explore the SMM. Furthermore, after the solution space is almost completely explored, the MC can be applied to quickly search for the few undiscovered solutions instead of checking boxes in the unexplored box buffer one by one. Once no unexplored node is specified by the MC, it is reasonable to early terminate the program.

Note that the output $\theta_s$ of \textsc{CalcNewSol} needs to be transformed into corresponding rotation matrices for \textsc{ExploreManif} to match the BnB search space. \textsc{SearchBuffer} contracts one of the unexplored boxes in the buffer $\mathcal{L}$ only when almost all solutions are found. Therefore, \textsc{SearchBuffer} is mainly used for eliminating infeasible regions rather than seeking for feasible solutions, which is necessary for the guaranteeing completeness.
%
% curve paving routine
For $1$ DoF SMMs, Fig.~\ref{fig: curve_paving} illustrates three theoretically possible patterns between the new solution $[\theta_{new}]$ and previously constructed curves. If no concatenation exists, a new curve containing $[\theta_{new}]$ will be created. If linked to one curve, $[\theta_{new}]$ will join it. If linked to two curves, $[\theta_{new}]$ will first join one of them and the curves will be merged into a longer one.

%% file: experiments.tex
\section{Numerical Experiments and Analysis}
\label{sec: experiments}
We evaluate performances of the proposed interval BnB-based IK solver for both non-redundant and redundant serial manipulators. The algorithm is implemented in C++, using the IBex library contracting interval boxes~\cite{chabert2009contractor} and the TRAC-IK library to provide the search heuristics~\cite{beeson2015trac}. All experiments are conducted on a desktop workstation with an $8$-core Genuine Intel i7-6700 CPU and \SI{15.6}{GB} memory.

\subsection{Solving IK for Non-Redundant Manipulators}
% In this experiment, we calculate IK solutions for the Universal Robot UR5\footnotemark ~(non-redundant $6$-DoF) using the vanilla BnB algorithm (without search heuristics) to examine the efficiency of the BnB mechanism.\footnotetext{\textbf{UR5:} http://www.universal-robots.com/products/ur5-robot/} Benefits of using search heuristics will be shown in the experiments on redundant manipulators. We assess the accuracy of joint solutions (\emph{i.e.}, $\textsc{Width}([\theta])$ and the computation time for $1000$ feasible EE poses generated by calculating forward kinematics from randomly sampled joint configurations.

In this experiment, we calculate IK solutions for the Universal Robot UR5 (non-redundant $6$-DoF) using the vanilla BnB algorithm (without search heuristics) to examine the efficiency of the BnB mechanism. Benefits of using search heuristics will be shown in the experiments on redundant manipulators. We assess the accuracy of joint solutions (\emph{i.e.}, $\textsc{Width}([\theta])$ and the computation time for $1000$ feasible EE poses generated by calculating forward kinematics from randomly sampled joint configurations.

\begin{figure}
    \centering
    \includegraphics[width=\linewidth]{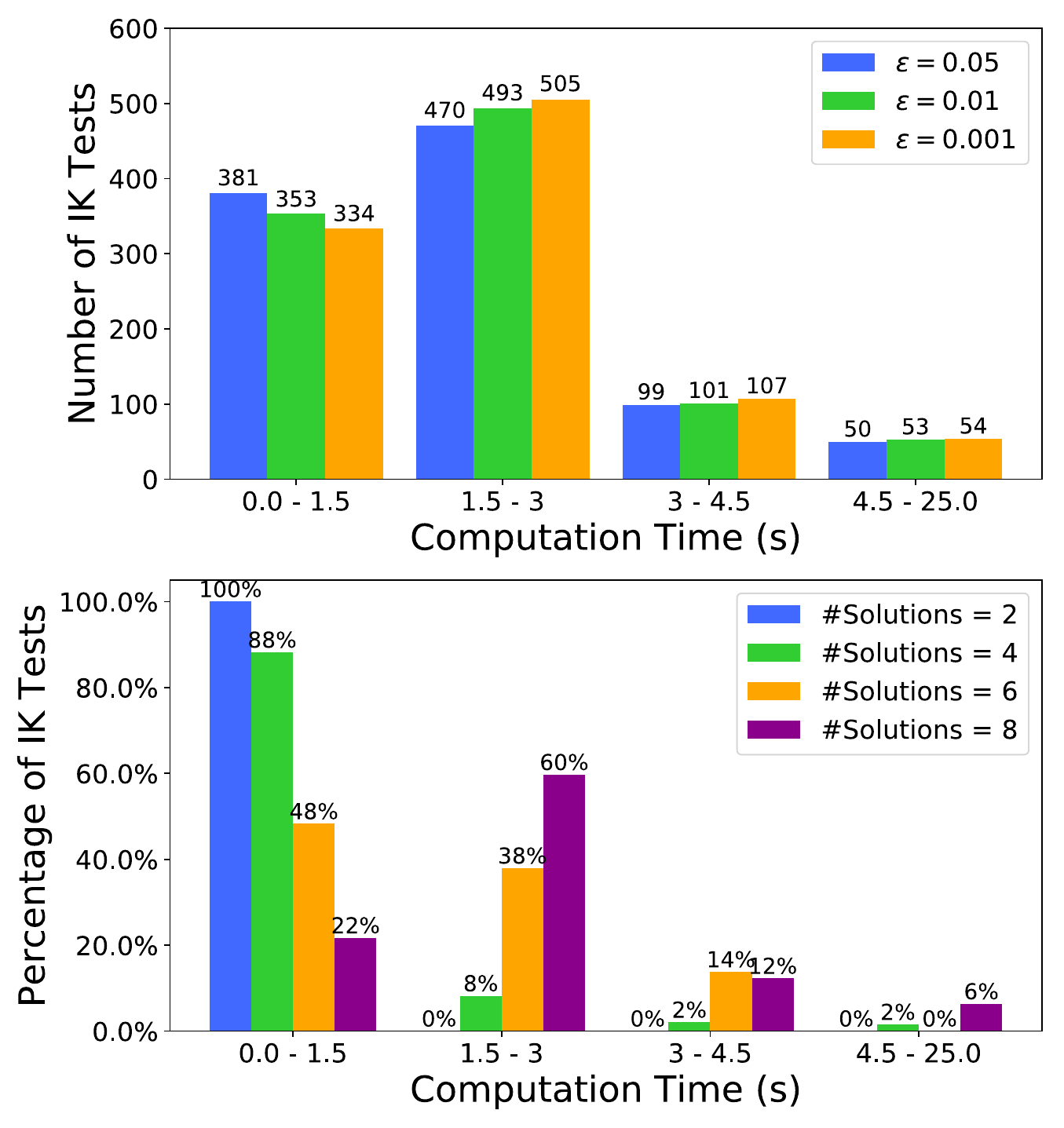}
    \caption{Computation time statistics. Top: Comparison of the computation time between different $\epsilon$. Bottom: Relationship between the computational performance of the vanilla BnB algorithm with $\epsilon = 0.05$ and the number of IK solutions.}
    \label{fig: bnb_time_stats}
    \vspace{-10pt}
\end{figure}

Since $\epsilon$ could theoretically affect both accuracy and efficiency, we test its impact on the IK solution by solving each problem with three different values of $\epsilon$: $0.001$, $0.01$ and $0.05$. As shown in Fig.~\ref{fig: bnb_time_stats}, for all values of $\epsilon$, over $80\%$ of the $1000$ IK computations finish within $3$ seconds. Besides, it appears that tests with larger $\epsilon$ tend to spend less time and vice versa. The average computation time and joint angle accuracy of the $1000$ tests are summarized in Table~\ref{tab: avg_time_acc}.

% \begin{figure}
%     \centering
%     \begin{subfigure}[b]{\linewidth}
%         \centering
%         \includegraphics[width=\linewidth]{figs/bnb_time_histogram_l.pdf}
%         \caption{}
%         % \caption{Comparison of the computation time between different $\epsilon$}
%         \label{fig: bnb_time_hist}
%     \end{subfigure}
%     \begin{subfigure}[b]{\linewidth}
%         \centering
%         \includegraphics[width=\textwidth]{figs/bnb_time_percent_l.pdf}
%         \caption{}
%         % \caption{Relationship between the computational performance of the vanilla BnB algorithm with $\epsilon = 0.05$ and the number of solutions to the IK problem.}
%         \label{fig: bnb_time_percent}
%     \end{subfigure}
%     \caption{Computation time profile of the vanilla BnB algorithm.}
% \end{figure}

% \begin{figure}[!htb]
%     \centering
%     \includegraphics[width=.8\linewidth]{figs/bnb_time_histogram.pdf}
%     \includegraphics[width=.8\linewidth]{figs/bnb_time_percent.pdf}
%     \caption{Computation time profile for the IK solution generation using the vanilla BnB algorithm.}
%     \label{fig: bnb_time_hist}
% \end{figure}
% On average, the vanilla BnB algorithm consumes around $2$ seconds for all tested $\epsilon$ as shown in Table~\ref{tab: avg_time_acc}. It suggests a moderate impact of $\epsilon$ on the algorithm's efficiency. In addition, Tabel~\ref{tab: avg_time_acc} also summarizes the average accuracy of joint solutions over the $1000$ tests.
\begin{table}
\centering
\caption{The average computation time and joint accuracy.}
\begin{tabular}{| c | c | c | c |}
\hline
$\epsilon$ & $0.05$ & $0.01$ & $0.001$ \\ \hline
Computation Time (ms) & $2022$ & $2097$ & $2131$ \\ \hline
Average Accuracy (rad) & $0.0033$ & $0.0001$ & $1.7921e^{-6}$\\ \hline
\end{tabular}
\label{tab: avg_time_acc}
\vspace{-10pt}
\end{table}

On the other hand, Fig.~\ref{fig: bnb_time_stats} shows that, with a fixed $\epsilon$, the computation time varies from less than \SI{1}{s} to \SI{25}{s}. This is legitimate because the efficiency of BnB algorithms depends significantly on the coefficient values of the IK equation as functions of the EE pose. Here, we investigate the influence of the number of solutions to the IK problem --- one of the indications of the equation's property --- on the algorithm's efficiency. The number of solutions is adopted according to the officially provided analytical IK solver. In fact, the vanilla BnB algorithm finds the same number of solutions. It is obvious that the more solutions a problem has, the more computation time is needed, even though all problems have the same search space (determined by joint limits) to be thoroughly explored. To explain this phenomenon, one should notice that the main factor in the efficiency of a BnB algorithm is the speed of deciding whether a region is infeasible. Early validation of infeasibility will nullify the need to explore sub-regions, thereby saving more time. In general, it is hard to make such validation for regions close to a solution, for example, in the extreme situation of cluster problem~\cite{schichl2004exclusion}. More solutions imply that more regions contiguous to them need to be patiently examined and thus more time-consuming.
% \begin{figure} [!htb]
%     \centering
%     \includegraphics[width=.8\linewidth]{figs/bnb_time_percent.pdf}
%     \caption{Relationship between the computational performance of the vanilla BnB algorithm without any heuristics and the number of solutions to the IK problem. Here we fix the parameter $\epsilon$ to $0.05$. The number of solutions is adopted according to the officially provided analytical IK solver. In fact, the vanilla BnB algorithm finds the same number of solutions.}
%     \label{fig: bnb_time_percent}
% \end{figure}

\subsection{Solving IK for Redundant Manipulators}
In this experiment, we characterize the SMMs for the KUKA LBR iiwa robot ($7$-DoF with $1$-DoF redundancy) using the BnB-based IK solver with search heuristics. We conduct experiments for $4$ different EE poses with $\epsilon = \{0.01, 0.05\}$ and two manifold exploration strategies (the DFS and the MC). According to the official brochure, ranges of motion for joints (from the first one to the last one) are respectively restricted to $\pm$\ang{170}, $\pm$\ang{120}, $\pm$\ang{170}, $\pm$\ang{120}, $\pm$\ang{170}, $\pm$\ang{120}, $\pm$\ang{175}.

\begin{table}
	\centering
	\caption{Results of experiments on KUKA iiwa.}
	\begin{tabular}{|c|c|c|c|c|c|}
		\hline
		\multirow{2}{*}{} & \multicolumn{2}{c|}{\begin{tabular}[c]{@{}c@{}}Early Termination\\ Time (s)\end{tabular}} & \multicolumn{2}{c|}{Termination Time (s)} & \multirow{2}{*}{\begin{tabular}[c]{@{}c@{}}Number of\\ SMMs\end{tabular}} 
		\\ \cline{2-5}
		& $\epsilon = 0.01$ & $\epsilon = 0.05$      & $\epsilon = 0.01$   & $\epsilon = 0.05$   &                                    \\ \hline
		Test 1 &  $1798.99$ & $606.23$ & $1997.07$ & $821.562$ & $23$ \\ \hline
		Test 2 &  $1841.22$ & $512.87$ & $2733.29$ & $1001.59$ & $21$ \\ \hline
		Test 3 &  $3048.80$ & $1015.05$ & $5076.63$ & $1994.76$ & $15$ \\ \hline
		Test 4 &  $4206.80$ & $1686.62$ & $6045.55$ & $3145.44$ & $37$ \\ \hline
	\end{tabular}
	\label{tab: redundant_expr}
    \vspace{-10pt}
\end{table}

\begin{figure*}[t]
	\centering
	\includegraphics[width=.99\linewidth]{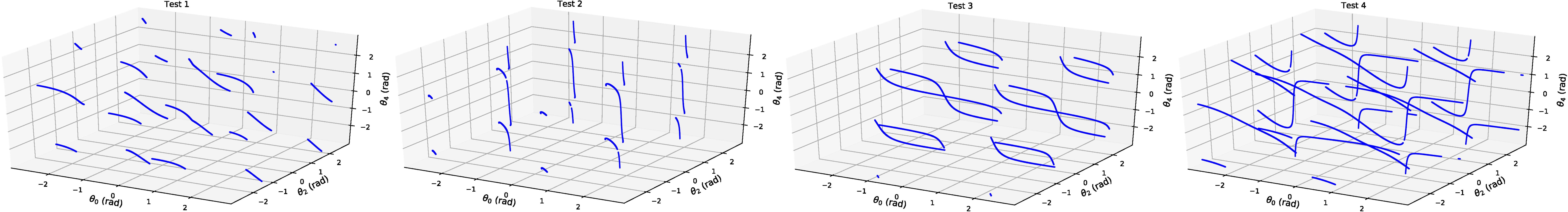}
	\caption{Self-motion manifolds projected onto $\theta_0$, $\theta_2$, and $\theta_4$. Note that the all the solutions are distinct. They just appear overlapping because they are projected onto the space with lower dimensions.}
	\label{fig: projection_3d}
\end{figure*}

\begin{figure}[t]
    \centering
    \begin{subfigure}[b]{\linewidth}
        \centering
        \includegraphics[width=\linewidth]{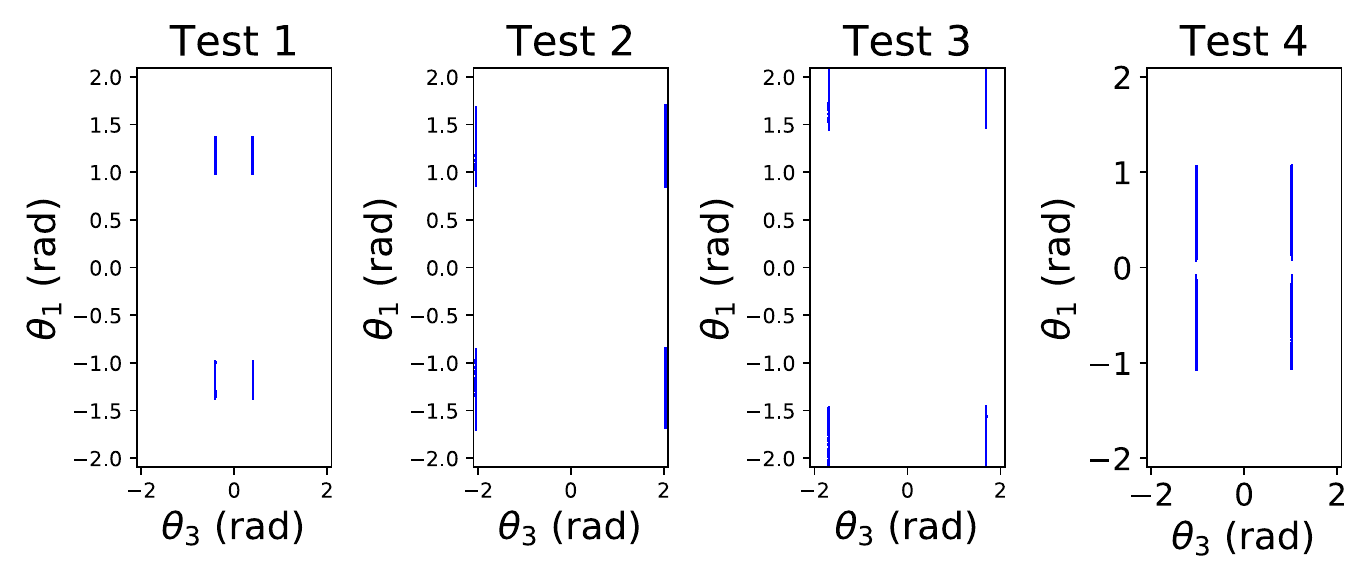}
        \vspace{-20pt}
        \caption{Particularity of $\theta_3$.}
        \label{fig: pecularity_jnt3}
    \end{subfigure}
    \begin{subfigure}[b]{\linewidth}
        \centering
        \includegraphics[width=\textwidth]{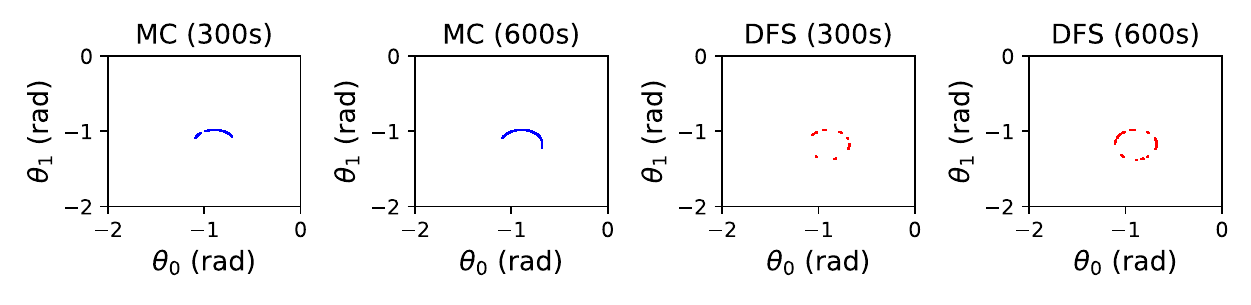}
        \vspace{-20pt}
        \caption{Manifold exploration by MC and DFS at $300$s, $600$s in Test 1.}
        \label{fig: manif_grow}
    \end{subfigure}
    \caption{2D projection of iiwa's SMMs}
\end{figure}

The results are listed in Table~\ref{tab: redundant_expr}, where early termination time represents the time at which the speed of finding a solution drastically reduces and the MC could be used to explore the remaining manifold around the end points of each paved curves; termination time stands for the time at which the entire space is searched. Like in the non-redundant cases, the computation time is pose dependent. Fig.~\ref{fig: projection_3d} depicts the projected SMMs onto $\theta_0$, $\theta_2$ and $\theta_4$. Curves in Test $4$ appear to be more convoluted, in which the IK solver spends the longest time. Note that joint limits could cut curves. Therefore, the numbers of SMMs in Test $1$, $2$ and $4$ exceed the theoretical value derived in~\cite{burdick1989inverse}.

\begin{figure}[t]
    \centering
    \includegraphics[width=\linewidth]{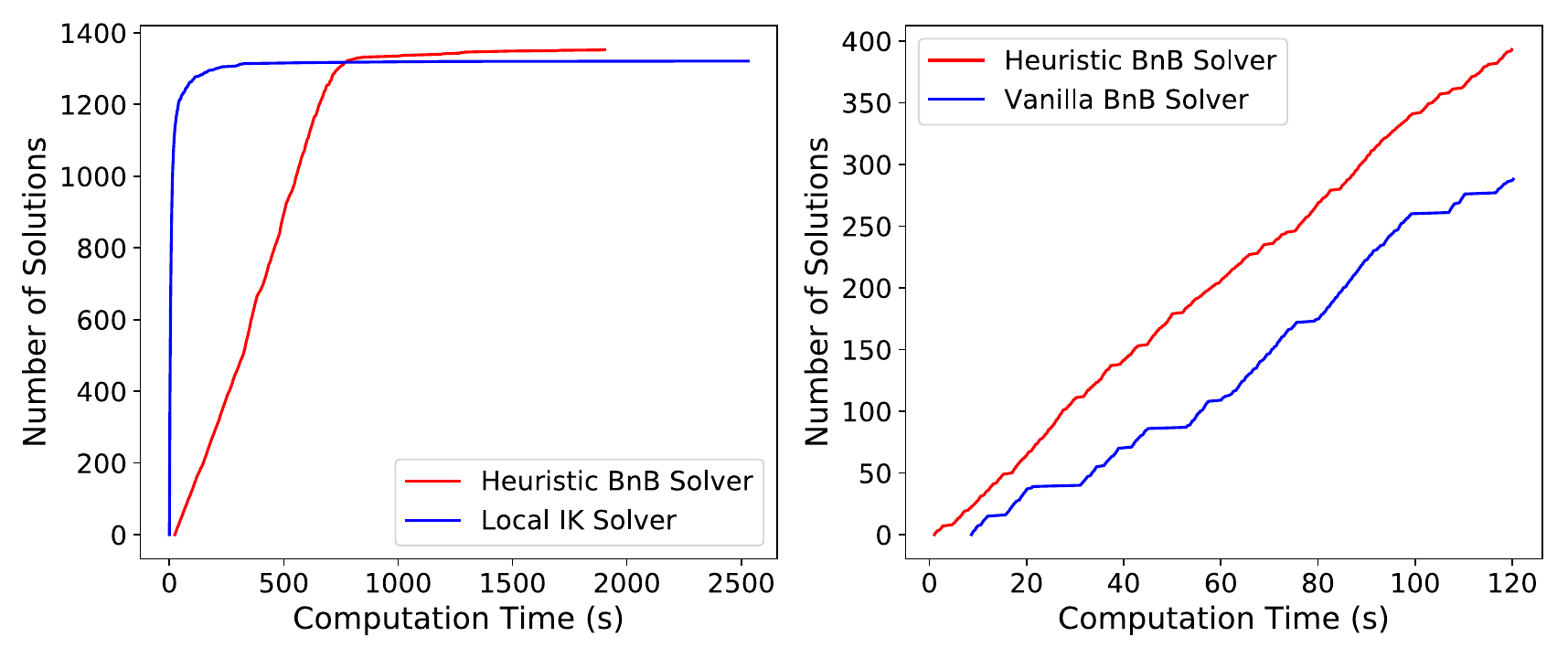}
    \caption{Left: Comparison of computation time of finding all IK solutions between the BnB solver with $\epsilon = 0.05$ and the instantaneous IK solver. Right: Comparison of computation time between heuristic BnB solver and vanilla BnB solver.}
    \label{fig: comp_bnb_combine}
\end{figure}

By observing the $2$D projection of the SMMs, as shown in Fig.~\ref{fig: pecularity_jnt3}, we find that for the iiwa robot, there are only two narrow interval values of $\theta_3$ for each EE pose. This feature throws light on the kinematic mechanism of iiwa, which may be further utilized in the derivation of its analytical IK algorithm. One possible strategy is, once the values of $\theta_3$ are figured out with geometric methods, the whole arm would be divided into two non-redundant sub-arms at $\theta_3$, for which the analytical IK equations may be easily developed. Continuity of manifold exploration with the MC and the DFS is compared in Fig.~\ref{fig: manif_grow}, which shows that the MC generates more continuous solutions while the DFS generates more diverse solutions.

% \begin{figure}[H]
% 	\centering
% 	\includegraphics[width=\linewidth]{figs/iiwa_special_jnt3_l.pdf}
% 	\caption{Particularity of $\theta_3$.}
% 	\label{fig: pecularity_jnt3}
% \end{figure}

For a comparison with the BnB algorithm, we test the computation time of finding all IK solutions for iiwa by repeatedly calling an instantaneous IK solver with uniformly sampled initial joint values. Fig.~\ref{fig: comp_bnb_combine} shows that the local IK solver spends longer time to obtain all solutions. Apart from being an incomplete algorithm, the more serious issue of this random sampling method is that it can not determine when to terminate the program. We also compare the efficiency of the heuristic BnB solver and the vanilla BnB solver as illustrated in Fig.~\ref{fig: comp_bnb_combine}. It is obvious that the heuristic BnB algorithm can find solutions without long pauses that occur in the vanilla solver.

% \begin{figure}[H]
%     \centering
%     \includegraphics[width=\linewidth]{figs/manif_grow_l.pdf}
%     \caption{Manifold exploration with MC and DFS for Test 1}
%     \label{fig: manif_grow}
% \end{figure}

% \begin{figure}
%     \centering
%     \begin{subfigure}[b]{.49\linewidth}
%         \centering
%         \includegraphics[width=.7\linewidth]{figs/comp_bnb_localik.pdf}
%         \caption{Comparison of computation time of finding all IK solutions between the BnB solver with $\epsilon = 0.05$ and the instantaneous IK solver.}
%         \label{fig: comp_bnb_local}
%     \end{subfigure}
%     \begin{subfigure}[b]{.49\linewidth}
%         \centering
%         \includegraphics[width=\textwidth]{figs/comp_heur_vanilla_l.pdf}
%         % \caption{Manifold exploration with MC and DFS for Test 1.}
%         \label{fig: comp_heur_vanilla}
%     \end{subfigure}
%     \caption{2D projection of iiwa's SMMs}
% \end{figure}

% \begin{figure}[H]
%     \centering
%     \includegraphics[width=.8\linewidth]{figs/comp_bnb_localik.pdf}
%     \caption{Comparison of computation time of finding all IK solutions between the BnB solver with $\epsilon = 0.05$ and the instantaneous IK solver.}
%     \label{fig: comp_bnb_local}
% \end{figure}

% \begin{figure}[H]
%     \centering
%     \includegraphics[width=.8\linewidth]{figs/comp_heur_vanilla.pdf}
%     \caption{Caption}
%     \label{fig: comp_heur_vanilla}
% \end{figure}

%% file: conclusion.tex
\section{Conclusion and Future Works}
\label{sec: conclusion}
This paper proposes an interval BnB-based algorithm with search heuristics provided by local search methods to completely solve the generalized IK problem for redundant serial manipulators. The proposed algorithm could also be utilized in anytime manner. In addition, it could generate patches of neighboring solutions that could potentially enable local reconfiguration of the robot. As an IK algorithm that guarantees completeness, it is reasonably efficient as is demonstrated by numerical experiments. Our future work may include parallel acceleration of the BnB solver.